\documentclass[14pt]{article}
\usepackage{amsmath, amssymb, amsthm}
\usepackage{graphicx}
\usepackage{algorithm}
\usepackage{algpseudocode}
\usepackage{booktabs}
\usepackage{hyperref}
\usepackage{geometry}
\usepackage{subcaption}
\theoremstyle{plain}
\newtheorem{theorem}{Theorem}
\newtheorem{proposition}{Proposition}

\title{Dynamic Spectral Backpropagation for Efficient Neural Network Training}
\author{
  Mannmohan Muthuraman \\
  Indian Institute of Technology, India \\
  Massachusetts Institute of Technology, USA \\
  \texttt{mannmohan\_muthurman@sloan.mit.edu}
}
\date{May 2025}

\begin{document}

\maketitle

\begin{abstract}
Dynamic Spectral Backpropagation (DSBP) enhances neural network training under resource constraints by projecting gradients onto principal eigenvectors, reducing complexity and promoting flat minima. Five extensions are proposed, dynamic spectral inference, spectral architecture optimization, spectral meta learning, spectral transfer regularization, and Lie algebra inspired dynamics, to address challenges in robustness, fewshot learning, and hardware efficiency. Supported by a third order stochastic differential equation (SDE) and a PAC Bayes limit, DSBP outperforms Sharpness Aware Minimization (SAM), Low Rank Adaptation (LoRA), and Model Agnostic Meta Learning (MAML) on CIFAR 10, Fashion MNIST, MedMNIST, and Tiny ImageNet, as demonstrated through extensive experiments and visualizations. Future work focuses on scalability, bias mitigation, and ethical considerations.
\end{abstract}

\section{Introduction}
\label{sec:intro}

Neural network training in resource constrained environments, such as limited datasets or computational budgets, often leads to overfitting and convergence to sharp minima with high Hessian eigenvalues \cite{Hochreiter1997a, Keskar2016a}. Standard backpropagation, which calculates full gradient updates, is computationally intensive and sensitive to the curvature of the loss landscape. Spectral methods \cite{Yoshida2018a} and sharpness aware optimization \cite{Foret2021a} suggest that exploiting the spectral properties of weight and activation matrices can enhance training efficiency and generalization by focusing updates on directions of maximal variance.

Dynamic Spectral Backpropagation (DSBP) is proposed, a training method that projects gradients onto the top eigenvectors of layer wise covariance matrices, reducing computational complexity from \( \mathcal{O}(d_l d_{l-1}) \) to \( \mathcal{O}(k d_l) \) and promoting flat minima. DSBP is extended through five innovative directions:
\begin{enumerate}
    \item Dynamic spectral inference for nonstationary data distributions.
    \item Spectral architecture optimization for computational efficiency.
    \item Spectral meta learning for fewshot learning tasks.
    \item Spectral transfer regularization for stable fine tuning.
    \item Lie algebra inspired dynamics for curvature aware optimization.
\end{enumerate}
Exploration of spectral properties began as an attempt to reduce training cost on edge hardware and revealed deeper connections to geometric and probabilistic principles in optimization.

A third order SDE and a PAC Bayes limit provide theoretical grounding. Experiments on CIFAR 10, Fashion MNIST, MedMNIST, and Tiny ImageNet demonstrate DSBP’s superiority over Sharpness Aware Minimization (SAM) \cite{Foret2021a}, Low Rank Adaptation (LoRA) \cite{Hu2021a}, and Model Agnostic Meta Learning (MAML) \cite{Finn2017a} in terms of accuracy, computational efficiency, and generalization. Visualizations provide insights into DSBP’s optimization dynamics.

The paper is structured as follows:
\begin{itemize}
    \item Section \ref{sec2} defines notation.
    \item Section \ref{sec3} presents DSBP.
    \item Section \ref{sec4} provides mathematical analysis.
    \item Section \ref{sec5} explores extensions.
    \item Section \ref{sec6} reports experimental results.
    \item Section \ref{sec7} outlines future developments.
    \item Section \ref{sec8} describes visualizations.
    \item Section \ref{sec9} concludes.
\end{itemize}

\section{Preliminaries}
\label{sec2}

A neural network with \( L \) layers is parameterized by weights \( W = \{W_l^{(t)}\}_{l=1}^L \), where \( W_l^{(t)} \in \mathbb{R}^{d_l \times d_{l-1}} \) is the weight matrix for layer \( l \) at iteration \( t \). The training dataset is drawn from a distribution \( \mathcal{D} \), with a training set \( \mathcal{S} \). For a mini batch \( \gamma \subset \mathcal{S} \), the loss function is denoted \( f_\gamma(W) \), with empirical loss \( f_{\mathcal{S}}(W) = \mathbb{E}_{\gamma \sim \mathcal{S}}[f_\gamma(W)] \) and true loss \( f_{\mathcal{D}}(W) = \mathbb{E}_{\gamma \sim \mathcal{D}}[f_\gamma(W)] \). The gradient is \( \nabla f_\gamma(W) \), the Hessian is \( \nabla^2 f_\gamma(W) \), and the \( k \)th order derivative is \( \nabla^k f_\gamma(W) \).

The weight covariance matrix for layer \( l \) is defined as:
\[
C_l^{(t)} = W_l^{(t)T} W_l^{(t)} \in \mathbb{R}^{d_{l-1} \times d_{l-1}},
\]
with eigenvectors \( \{e_{l,i}^{(t)}\}_{i=1}^{d_{l-1}} \), eigenvalues \( \{\lambda_{l,i}^{(t)}\}_{i=1}^{d_{l-1}} \), ordered such that \( \lambda_{l,1}^{(t)} \geq \lambda_{l,2}^{(t)} \geq \cdots \), and top eigenvector \( v_{l,1}^{(t)} = e_{l,1}^{(t)} \). For activations \( A_l^{(t)} \in \mathbb{R}^{n \times d_l} \), where \( n \) is the batch size, the activation covariance is:
\[
C_l^{(t)} = A_l^{(t)T} A_l^{(t)} \in \mathbb{R}^{d_l \times d_l}.
\]
The Euclidean norm is denoted by \( \|\cdot\| \), and the top \( k \) eigenvector subspace is:
\[
\mathcal{V}_l^k = \text{span}\{e_{l,1}^{(t)}, \ldots, e_{l,k}^{(t)}\}.
\]
The projection matrix onto \( \mathcal{V}_l^k \) is:
\[
P_{\mathcal{V}_l^k} = \sum_{i=1}^k e_{l,i}^{(t)} (e_{l,i}^{(t)})^T.
\]

Standard backpropagation updates weights as:
\[
W_{t+1} = W_t - \eta \nabla f_\gamma(W_t),
\]
where \( \eta \) is the learning rate. Eigenvectors of the covariance matrices capture directions of maximum variance, which DSBP leverages to improve training efficiency.

\section{Dynamic Spectral Backpropagation}
\label{sec3}

\subsection{Methodology}
DSBP enhances training efficiency by focusing weight updates on the most impactful directions in the network’s weight and activation spaces. For each layer \( l \), the method operates as follows:
\begin{enumerate}
    \item Calculate the activation covariance matrix:
       \[
       C_l^{(t)} = A_l^{(t)T} A_l^{(t)} \in \mathbb{R}^{d_l \times d_l},
       \]
       where \( A_l^{(t)} \in \mathbb{R}^{n \times d_l} \) represents the activations for a mini batch of size \( n \).
    \item Estimate the top \( k \) eigenvectors \( \{e_{l,i}^{(t)}\}_{i=1}^k \) of \( C_l^{(t)} \) using the power iteration method (5 iterations per eigenvector).
    \item Project the gradient onto the subspace spanned by these eigenvectors:
       \[
       \tilde{\nabla} f_\gamma(W_l^{(t)}) = \sum_{i=1}^k \langle \nabla f_\gamma(W_l^{(t)}), e_{l,i}^{(t)} \rangle e_{l,i}^{(t)},
       \]
       where \( \nabla f_\gamma(W_l^{(t)}) \in \mathbb{R}^{d_l \times d_{l-1}} \) is the gradient of the loss with respect to the layer’s weights.
    \item Update the weights with a sharpness regularization term:
       \[
       W_{l,t+1} = W_l^{(t)} - \eta \tilde{\nabla} f_\gamma(W_l^{(t)}) - \beta \lambda_{l,1}^{(t)} e_{l,1}^{(t)} (e_{l,1}^{(t)})^T,
       \]
       where \( \beta \geq 0 \) is a regularization parameter, and \( \lambda_{l,1}^{(t)} \) is the largest eigenvalue of \( C_l^{(t)} \).
\end{enumerate}

By taking the approach this way, the gradient’s dimensionality is reduced from \( d_l \times d_{l-1} \) to \( k \), achieving a computational complexity of \( \mathcal{O}(k d_l) \). The sharpness regularization term \( \beta \lambda_{l,1}^{(t)} e_{l,1}^{(t)} (e_{l,1}^{(t)})^T \) penalizes updates along directions of high curvature, promoting flatter minima that enhance generalization.

\subsection{Algorithm}

\begin{algorithm}
\caption{Dynamic Spectral Backpropagation (DSBP)}
\label{alg:dsbp}
\begin{algorithmic}[1]
\State \textbf{Input}: Weights \( W = \{W_l\}_{l=1}^L \), dataset \( \mathcal{S} \), learning rate \( \eta \), projection dimension \( k \), update interval \( p \), pruning threshold \( \tau_0 \), regularization strength \( \beta \), total iterations \( T \).
\State \textbf{Output}: Trained weights \( W \).
\For{\( t = 1, 2, \ldots, T \)}
    \State Sample mini batch \( \gamma_t \subset \mathcal{S} \).
    \For{each layer \( l = 1, 2, \ldots, L \)}
        \If{\( t \mod p = 0 \)}
            \State Calculate activation covariance: \( C_l^{(t)} = A_l^{(t)T} A_l^{(t)} \).
            \State Estimate top \( k \) eigenvectors \( \{e_{l,i}^{(t)}\}_{i=1}^k \) and eigenvalues \( \{\lambda_{l,i}^{(t)}\}_{i=1}^k \) using power iteration (5 iterations).
            \State Calculate dynamic pruning threshold: \( \tau_t = \tau_0 \exp(-\beta t/T) \).
            \State Prune weights: \( W_l^{(t)} \leftarrow W_l^{(t)} \cdot \mathbb{I}(|\langle W_l^{(t)}, e_{l,i}^{(t)} \rangle| \geq \tau_t) \).
        \EndIf
        \State Calculate gradient: \( \nabla f_{\gamma_t}(W_l^{(t)}) \).
        \State Project gradient: \( \tilde{\nabla} f_{\gamma_t}(W_l^{(t)}) = \sum_{i=1}^k \langle \nabla f_{\gamma_t}(W_l^{(t)}), e_{l,i}^{(t)} \rangle e_{l,i}^{(t)} \).
        \State Update weights: \( W_{l,t+1} = W_l^{(t)} - \eta \tilde{\nabla} f_{\gamma_t}(W_l^{(t)}) - \beta \lambda_{l,1}^{(t)} e_{l,1}^{(t)} (e_{l,1}^{(t)})^T \).
    \EndFor
\EndFor
\State \Return \( W \).
\end{algorithmic}
\end{algorithm}

\section{Mathematical Analysis}
\label{sec4}

\subsection{Third Order Stochastic Differential Equation}
\label{subsec:sde}

DSBP’s optimization dynamics are modeled using a third order stochastic differential equation (SDE):
\[
dX_t = -\nabla \tilde{f}^{\text{DSBP}}(X_t) dt + \sqrt{\eta} \left( \Sigma^{\text{DSBP}}(X_t) \right)^{\frac{1}{2}} dW_t,
\]
where:
\begin{itemize}
    \item \( X_t \in \mathbb{R}^d \), \( d = \sum_l d_l d_{l-1} \), represents the weights at time \( t \).
    \item \( \tilde{f}^{\text{DSBP}}(X_t) = f(X_t) + \beta \sum_l \mathbb{E}[\lambda_{l,1}(\nabla^2 f_\gamma(X_t))] \) is the modified loss, incorporating a sharpness regularization term.
    \item \( \Sigma^{\text{DSBP}}(X_t) = \mathbb{E}[(\tilde{\nabla} f_\gamma - \nabla f)^T (\tilde{\nabla} f_\gamma - \nabla f)] \) is the covariance of the projection error.
    \item \( W_t \) is a standard Brownian motion, modeling mini batch stochasticity.
    \item \( \eta > 0 \) is the learning rate.
\end{itemize}

\textbf{Derivation}: For layer \( l \), the update is:
\[
W_{l,t+1} = W_l^{(t)} - \eta P_{\mathcal{V}_l^k} \nabla f_{\gamma_t}(W_l^{(t)}) - \beta \lambda_{l,1}^{(t)} e_{l,1}^{(t)} (e_{l,1}^{(t)})^T,
\]
where \( \tilde{\nabla} f_{\gamma_t} = P_{\mathcal{V}_l^k} \nabla f_{\gamma_t} \), and \( P_{\mathcal{V}_l^k} = \sum_{i=1}^k e_{l,i}^{(t)} (e_{l,i}^{(t)})^T \). The update increment is:
\[
\Delta W_l^{(t)} = W_{l,t+1} - W_l^{(t)} = -\eta P_{\mathcal{V}_l^k} \nabla f_{\gamma_t}(W_l^{(t)}) - \beta \lambda_{l,1}^{(t)} e_{l,1}^{(t)} (e_{l,1}^{(t)})^T.
\]
Expanding the loss at the updated weights:
\[
f_{\gamma_t}(W_{l,t+1}) = f_{\gamma_t}(W_l^{(t)} + \Delta W_l^{(t)}).
\]
Applying a third order Taylor expansion:
\[
\begin{aligned}
f_{\gamma_t}(W_l^{(t)} + \Delta W_l^{(t)}) \approx f_{\gamma_t}(W_l^{(t)}) + \text{tr}(\nabla f_{\gamma_t}^T \Delta W_l^{(t)}) + \frac{1}{2} \text{tr}((\Delta W_l^{(t)})^T \nabla^2 f_{\gamma_t} \Delta W_l^{(t)}) + \frac{1}{6} \nabla^3 f_{\gamma_t} (\Delta W_l^{(t)}, \Delta W_l^{(t)}, \Delta W_l^{(t)}),
\end{aligned}
\]
where \( \text{tr}(\cdot) \) denotes the trace for matrix arguments, and \( \nabla^3 f_{\gamma_t} \) is a trilinear form. Calculate each term:

- \textbf{First Order Term}:
  \[
  \begin{aligned}
  \text{tr}(\nabla f_{\gamma_t}^T \Delta W_l^{(t)}) = -\eta \text{tr}(\nabla f_{\gamma_t}^T P_{\mathcal{V}_l^k} \nabla f_{\gamma_t}) - \beta \lambda_{l,1}^{(t)} \text{tr}(\nabla f_{\gamma_t}^T e_{l,1}^{(t)} (e_{l,1}^{(t)})^T).
  \end{aligned}
  \]
  Streamlining this by noting \( \text{tr}(\nabla f_{\gamma_t}^T P_{\mathcal{V}_l^k} \nabla f_{\gamma_t}) = \sum_{i=1}^k (\nabla f_{\gamma_t}^T e_{l,i}^{(t)})^2 \), and \( \text{tr}(\nabla f_{\gamma_t}^T e_{l,1}^{(t)} (e_{l,1}^{(t)})^T) = (e_{l,1}^{(t)})^T \nabla f_{\gamma_t} \), the expression becomes:
  \[
  -\eta \sum_{i=1}^k (\nabla f_{\gamma_t}^T e_{l,i}^{(t)})^2 - \beta \lambda_{l,1}^{(t)} (e_{l,1}^{(t)})^T \nabla f_{\gamma_t}.
  \]

- \textbf{Second Order Term}:
  \[
  \begin{aligned}
  &\text{tr}((\Delta W_l^{(t)})^T \nabla^2 f_{\gamma_t} \Delta W_l^{(t)}) \\
  &= \text{tr}\big( \big( -\eta P_{\mathcal{V}_l^k} \nabla f_{\gamma_t} - \beta \lambda_{l,1}^{(t)} e_{l,1}^{(t)} (e_{l,1}^{(t)})^T \big)^T \nabla^2 f_{\gamma_t} \big( -\eta P_{\mathcal{V}_l^k} \nabla f_{\gamma_t} - \beta \lambda_{l,1}^{(t)} e_{l,1}^{(t)} (e_{l,1}^{(t)})^T \big) \big).
  \end{aligned}
  \]
  Expanding:
  \[
  \begin{aligned}
  &= \eta^2 \text{tr}( (P_{\mathcal{V}_l^k} \nabla f_{\gamma_t})^T \nabla^2 f_{\gamma_t} P_{\mathcal{V}_l^k} \nabla f_{\gamma_t} ) + 2 \eta \beta \lambda_{l,1}^{(t)} \text{tr}( (P_{\mathcal{V}_l^k} \nabla f_{\gamma_t})^T \nabla^2 f_{\gamma_t} e_{l,1}^{(t)} (e_{l,1}^{(t)})^T ) \\
  &\quad + \beta^2 (\lambda_{l,1}^{(t)})^2 \text{tr}( (e_{l,1}^{(t)} (e_{l,1}^{(t)})^T)^T \nabla^2 f_{\gamma_t} e_{l,1}^{(t)} (e_{l,1}^{(t)})^T ).
  \end{aligned}
  \]
  Streamlining:
  \[
  \begin{aligned}
  &= \eta^2 \sum_{i,j=1}^k (e_{l,i}^{(t)})^T \nabla f_{\gamma_t} (e_{l,j}^{(t)})^T \nabla^2 f_{\gamma_t} e_{l,j}^{(t)} (e_{l,i}^{(t)})^T \nabla f_{\gamma_t} \\
  &\quad + 2 \eta \beta \lambda_{l,1}^{(t)} \sum_{i=1}^k (e_{l,i}^{(t)})^T \nabla f_{\gamma_t} (e_{l,1}^{(t)})^T \nabla^2 f_{\gamma_t} e_{l,1}^{(t)} + \beta^2 (\lambda_{l,1}^{(t)})^2 (e_{l,1}^{(t)})^T \nabla^2 f_{\gamma_t} e_{l,1}^{(t)}.
  \end{aligned}
  \]

- \textbf{Third Order Term}:
  \[
  \begin{aligned}
  &\nabla^3 f_{\gamma_t} (\Delta W_l^{(t)}, \Delta W_l^{(t)}, \Delta W_l^{(t)}) \\
  &\approx -\eta^3 \sum_{i,j,k=1}^k (e_{l,i}^{(t)})^T \nabla f_{\gamma_t} (e_{l,j}^{(t)})^T \nabla f_{\gamma_t} (e_{l,k}^{(t)})^T \nabla^3 f_{\gamma_t} (e_{l,i}^{(t)}, e_{l,j}^{(t)}, e_{l,k}^{(t)}) \\
  &\quad - 3 \eta^2 \beta \lambda_{l,1}^{(t)} \sum_{i,j=1}^k (e_{l,i}^{(t)})^T \nabla f_{\gamma_t} (e_{l,j}^{(t)})^T \nabla f_{\gamma_t} (e_{l,1}^{(t)})^T \nabla^3 f_{\gamma_t} (e_{l,i}^{(t)}, e_{l,j}^{(t)}, e_{l,1}^{(t)}).
  \end{aligned}
  \]

\textbf{Expectation Over Mini Batches}: Take expectations:
\[
\mathbb{E}_{\gamma_t} [\nabla f_{\gamma_t}^T P_{\mathcal{V}_l^k} \nabla f_{\gamma_t}] = \nabla f^T P_{\mathcal{V}_l^k} \nabla f,
\]
\[
\mathbb{E}_{\gamma_t} [(P_{\mathcal{V}_l^k} \nabla f_{\gamma_t})^T \nabla^2 f_{\gamma_t} P_{\mathcal{V}_l^k} \nabla f_{\gamma_t}] \approx \nabla f^T P_{\mathcal{V}_l^k} \mathbb{E}[\nabla^2 f_{\gamma_t}] P_{\mathcal{V}_l^k} \nabla f,
\]
\[
\mathbb{E}_{\gamma_t} [\nabla^3 f_{\gamma_t}] \approx \nabla (\nabla f^T \mathbb{E}[\nabla^2 f_{\gamma_t}] \nabla f).
\]
The SDE drift is:
\[
-\nabla \tilde{f}^{\text{DSBP}} = -\nabla f - \beta \sum_l \nabla \lambda_{l,1}(\nabla^2 f_\gamma), \quad \nabla \lambda_{l,1} \approx e_{l,1}^T \nabla^2 f_\gamma e_{l,1}.
\]
As \( \eta \to 0 \), the discrete updates converge to the SDE, with the third order term enhancing sharpness control.

The use of a third order SDE is inspired by prior work on higher order stochastic approximations in optimization \cite{Li2017a, Wibisono2016a}. To demonstrate its practical benefit, DSBP was compared with and without the third order term on CIFAR 10 using ResNet18. Including the third order term reduced the top Hessian eigenvalue from 0.8 to 0.5 over 100 epochs, leading to a 0.5\% accuracy improvement (96.3\% vs. 95.8\%) and faster convergence (170s vs. 185s per epoch), highlighting its role in stabilizing training.

\begin{proposition}[Order 1 Approximation]
Under Lipschitz gradients and bounded third derivatives, the SDE is an order 1 weak approximation, error \( \mathcal{O}(\eta) \).
\end{proposition}

\subsection{Generalization Limit}
\label{subsec:bound}

\begin{theorem}[Generalization]
For loss \( f \leq L \), third derivatives \( \leq C \), with probability \( 1-\delta \):
\[
\begin{aligned}
f_{\mathcal{D}}(W) \leq f_{\mathcal{S}}(W) + \frac{d \sigma^2}{2} \sum_l \lambda_{l,1}(\nabla^2 f_{\mathcal{S}}) + \frac{C d^3 \sigma^3}{6} + \frac{L}{2 \sqrt{n}} \sqrt{d \log \left(1 + \frac{\|W\|^2}{d \sigma^2}\right) + 2 \log \frac{1}{\delta}}.
\end{aligned}
\]
\end{theorem}

\textbf{Derivation}: Using the PAC Bayes framework \cite{Alquier2016a}, a posterior \( Q = \mathcal{N}(W, \sigma^2 I_d) \) and prior \( P = \mathcal{N}(0, \sigma_P^2 I_d) \) are defined. The limit is:
\[
f_{\mathcal{D}}(Q) \leq f_{\mathcal{S}}(Q) + \frac{1}{\beta} \left[ \text{KL}(Q \| P) + \log \frac{1}{\delta} + \Psi(\beta, n) \right].
\]
Set a limit for \( \Psi \):
\[
\Psi(\beta, n) \leq \frac{\beta^2 L^2}{8n}.
\]
Adjusting \( \beta \):
\[
\beta = \frac{\sqrt{8n (\text{KL}(Q \| P) + \log \frac{1}{\delta})}}{L}.
\]
Calculating KL:
\[
\text{KL}(Q \| P) \leq \frac{d}{2} \log \left(1 + \frac{\|W\|^2}{d \sigma^2}\right).
\]
Expected loss:
\[
\mathbb{E}_{\epsilon \sim \mathcal{N}(0, \sigma^2 I_d)} [f_{\mathcal{S}}(W + \epsilon)] \approx f_{\mathcal{S}}(W) + \frac{\sigma^2}{2} \sum_l \lambda_{l,1}(\nabla^2 f_{\mathcal{S}}) + \frac{C d^3 \sigma^3}{6}.
\]
Combine terms to obtain the limit.

\section{Advanced Extensions}
\label{sec5}

During experiments, challenges in handling nonstationary data and improving computational efficiency were noticed, motivating the development of these five extensions to DSBP. They are grouped by theme: robustness, fewshot learning, and hardware efficiency, providing detailed explanations, theoretical foundations, and practical examples.

\subsection{Robustness to Data and Model Variations}

\textbf{Dynamic Spectral Inference}: This extension was designed to adapt DSBP to nonstationary data distributions by dynamically adjusting the frequency of eigenvector recalculation. The update interval is:
\[
p_t = \frac{p_0}{1 + \alpha \text{Var}(\lambda_{l,1}^{(t)})},
\]
where \( p_0 = 100 \), \( \alpha = 0.1 \), and \( \text{Var}(\lambda_{l,1}^{(t)}) \) is the variance of the top eigenvalue over a sliding window of 10 iterations.

\textbf{Implementation Details}: Variance is calculated as:
\[
\text{Var}(\lambda_{l,1}^{(t)}) = \frac{1}{10} \sum_{s=t-9}^t \left( \lambda_{l,1}^{(s)} - \bar{\lambda}_{l,1}^{(t)} \right)^2,
\]
where \( \bar{\lambda}_{l,1}^{(t)} = \frac{1}{10} \sum_{s=t-9}^t \lambda_{l,1}^{(s)} \). Higher variance reduces \( p_t \), increasing update frequency. On MedMNIST, setting \( \alpha = 0.5 \) improved accuracy by 1.2\% over a static \( p = 100 \).

\textbf{Theoretical Basis}: The schedule minimizes projection error:
\[
\mathbb{E} [\| \tilde{\nabla} f_\gamma - \nabla f \|_F^2] \propto \sum_{i=k+1}^{d_l} \lambda_{l,i}^{(t)},
\]
ensuring alignment in nonstationary settings.

\textbf{Practical Example}: On MedMNIST, dynamic spectral inference outperformed SAM by a margin of 4.2\%, achieving 78.7\% accuracy, demonstrating its adaptability to data shifts.

\textbf{Spectral Transfer Regularization}: This was developed to stabilize fine tuning by aligning eigenvectors:
\[
\mathcal{L}_{\text{align}} = \sum_l \|e_{l,1}^{\text{pre}} - e_{l,1}^{\text{fine}}\|_2^2.
\]

\textbf{Implementation Details}: Alignment extends to the top \( m \) eigenvectors (\( m = 5 \)):
\[
\mathcal{L}_{\text{align}} = \sum_l \sum_{i=1}^m w_i \|e_{l,i}^{\text{pre}} - e_{l,i}^{\text{fine}}\|_2^2, \quad w_i = \frac{\lambda_{l,i}^{(0)}}{\sum_j \lambda_{l,j}^{(0)}}.
\]
Layer importance scales as \( \alpha_l = l/L \), with gradient clipping (\( \theta = 1 \)). On CIFAR 10, fine tuning ResNet18 retained 95.0\% accuracy, compared to 93.5\% without alignment.

\textbf{Theoretical Basis}: Alignment minimizes:
\[
\mathbb{E} [\| \nabla f_{\text{fine}} - \nabla f_{\text{pre}} \|_F^2],
\]
reducing catastrophic forgetting.

\textbf{Practical Example}: Fine tuning on Tiny ImageNet preserved pretrained features, achieving 65.4\% accuracy, a 1.6\% improvement over LoRA.

\subsection{Fewshot Learning}

\textbf{Spectral Meta Learning}: DSBP was tailored for fewshot learning by adjusting a spectral initialization across tasks:
\[
\min_W \sum_{\mathcal{T}} \mathbb{E}_{\gamma \sim \mathcal{T}} \left[ f_\gamma(W) + \beta \lambda_{1}(\nabla^2 f_\gamma(W)) \right].
\]

\textbf{Implementation Details}: A spectral memory buffer updates eigenvectors:
\[
e_{l,i}^{(\mathcal{T})} \leftarrow 0.9 e_{l,i}^{(\mathcal{T})} + 0.1 e_{l,i}^{(t)}.
\]
Task similarity uses cosine similarity:
\[
\text{sim}(\mathcal{T}_i, \mathcal{T}_j) = \frac{1}{k} \sum_{m=1}^k (e_{l,m}^{(\mathcal{T}_i)})^T e_{l,m}^{(\mathcal{T}_j)}.
\]
Regularization stabilizes initialization:
\[
\mathcal{L}_{\text{init}} = 0.01 \|W - W_{\text{pre}}\|_F^2.
\]
On MedMNIST (5 shot), this outperformed MAML, achieving 78.7\% accuracy compared to 73.2\%.

\textbf{Theoretical Basis}: The buffer minimizes:
\[
\mathbb{E}_{\mathcal{T}} [\| \tilde{\nabla} f_\gamma - \nabla f_\gamma \|_F^2],
\]
aligning initializations with task specific directions.

\textbf{Practical Example}: In a 5 shot MedMNIST task, DSBP adapted to new medical classes with minimal data, leveraging past eigenvector patterns.

\subsection{Hardware Efficiency and Optimization Stability}

\textbf{Spectral Architecture Optimization}: This was introduced to prune weights for computational efficiency:
\[
W_l^{(t)} \leftarrow W_l^{(t)} \cdot \mathbb{I}(|\langle W_l^{(t)}, e_{l,i}^{(t)} \rangle| \geq \tau_t),
\]
\[
\tau_t = \tau_0 \exp(-\beta t/T), \quad \tau_0 = 0.01, \quad \beta = 0.1.
\]

\textbf{Implementation Details}: Pruning occurs every 100 iterations with layerspecific thresholds \( \tau_{l,0} = \tau_0 \cdot \frac{\lambda_{l,1}^{(0)}}{\max_m \lambda_{m,1}^{(0)}} \). Sparsity targets are 50\% for convolutional layers and 30\% for fully connected layers. Reconstruction error is constrained:
\[
\|W_l^{(t)} - \tilde{W}_l^{(t)}\|_F^2 \leq 0.05 \sum_{i=1}^{d_l} \lambda_{l,i}^{(t)}.
\]
On CIFAR 10 with ResNet18, this reduced training time by 35\% (170s vs. 260s per epoch), maintaining 92.8\% accuracy.

\textbf{Theoretical Basis}: Pruning approximates the weight matrix in the top \( k \) subspace:
\[
\|W_l^{(t)} - \sum_{i=1}^k \langle W_l^{(t)}, e_{l,i}^{(t)} \rangle e_{l,i}^{(t)}\|_F^2,
\]
preserving expressive power.

\textbf{Practical Example}: On Fashion MNIST, pruning enabled efficient training of SimpleCNN, achieving 92.6\% accuracy with a 35\% reduction in training time.

\textbf{Lie Algebra Inspired Dynamics}: Updates were modeled as manifold flows to enhance optimization stability:
\[
W_{l,t+1} = \exp\left(-\eta \sum_i \lambda_{l,i} [e_{l,i}, \cdot]\right) W_{l,t},
\]
where \( [e_{l,i}, \cdot] \) is the Lie bracket.

\textbf{Implementation Details}: The Lie bracket is approximated:
\[
[e_{l,i}, W_l^{(t)}] \approx \frac{e_{l,i} (W_l^{(t)})^T - W_l^{(t)} e_{l,i}^T}{10^{-4}}.
\]
A 4th order Runge Kutta method ensures stability. Layers with similar \( \lambda_{l,1}^{(t)} \) are grouped. On CIFAR 10, this improved convergence speed by 10\%.

\textbf{Theoretical Basis}: Lie updates minimize:
\[
\text{tr}(W_l^{(t)T} \nabla^2 f_{\gamma_t} W_l^{(t)}),
\]
enhancing smoothness.

\textbf{Practical Example}: On Fashion MNIST, Lie dynamics reduced training oscillations, achieving 93.8\% accuracy, surpassing SGD’s 92.5\%.

\section{Experimental Validation}
\label{sec6}

\subsection{Setup}
Experiments were conducted to evaluate DSBP’s effectiveness across various datasets and hardware configurations:
\begin{itemize}
    \item \textbf{Datasets}: CIFAR 10 (50,000 images, 10 classes), Fashion MNIST (60,000 images, 10 classes), MedMNIST (1,000 images, 5 classes, fewshot medical imaging dataset), Tiny ImageNet (100,000 images, 200 classes).
    \item \textbf{Models}: ResNet18 (11.7M parameters), SimpleCNN (6 layer, 1M parameters), ViT S, Pruned DSBP (50\% weights removed via spectral pruning).
    \item \textbf{Baselines}: Stochastic Gradient Descent (SGD), Adam, Sharpness Aware Minimization (SAM) \cite{Foret2021a}, Low Rank Adaptation (LoRA) \cite{Hu2021a}, Model Agnostic Meta Learning (MAML) \cite{Finn2017a}.
    \item \textbf{Metrics}: Test accuracy (\%), training time (seconds).
    \item \textbf{Hardware}: NVIDIA RTX 4090 (lab environment), Google Cloud TPU.
    \item \textbf{Hyperparameters}: Learning rate \( \eta = 0.01 \), projection dimension \( k = 10 \), eigenvector update interval \( p = 100 \), pruning threshold \( \tau_0 = 0.01 \), regularization strength \( \beta = 0.1 \). Hyperparameters were tuned on a 10\% validation split, with three runs per experiment to report mean and standard deviation.
\end{itemize}

\subsection{Numerical Simulation}
To validate the SDE model, DSBP was simulated on a 2 layer MLP (784 100 10) trained on MNIST with \( \eta = 0.01 \) and \( k = 5 \). The training loss, test accuracy, and top Hessian eigenvalue were tracked over epochs. DSBP’s loss curve closely matched the discrete updates, with a lower approximation error than SGD, confirming the SDE’s accuracy in modeling its dynamics.

\subsection{Results}
DSBP outperformed baselines across all datasets:
\begin{itemize}
    \item \textbf{CIFAR 10}: DSBP achieved 96.3\% \(\pm\) 0.1\% accuracy on ResNet18, surpassing SAM (95.5\% \(\pm\) 0.1\%) and SGD (94.7\% \(\pm\) 0.2\%). For SimpleCNN, DSBP reached 92.6\% \(\pm\) 0.2\%, compared to SAM’s 91.8\% \(\pm\) 0.2\%.
    \item \textbf{Fashion MNIST}: DSBP achieved 93.8\% \(\pm\) 0.1\% accuracy, outperforming SAM (93.2\% \(\pm\) 0.1\%) and SGD (92.5\% \(\pm\) 0.2\%) by notable margins.
    \item \textbf{MedMNIST (5 shot)}: DSBP’s spectral meta learning yielded 78.7\% \(\pm\) 0.3\% accuracy, significantly better than MAML (73.2\% \(\pm\) 0.4\%) and SAM (74.5\% \(\pm\) 0.4\%).
    \item \textbf{Tiny ImageNet}: Accuracy improvements were observed, with DSBP at 65.4\% \(\pm\) 0.3\%, compared to SAM (64.1\% \(\pm\) 0.3\%) and LoRA (63.8\% \(\pm\) 0.4\%).
\end{itemize}

\begin{table}[t]
\centering
\caption{Test accuracy (\%).}
\begin{tabular}{lcccc}
\toprule
Method & CIFAR 10 & Fashion MNIST & MedMNIST & Tiny ImageNet \\
\midrule
DSBP & 96.3 $\pm$ 0.1 & 93.8 $\pm$ 0.1 & 78.7 $\pm$ 0.3 & 65.4 $\pm$ 0.3 \\
SAM & 95.5 $\pm$ 0.1 & 93.2 $\pm$ 0.1 & 74.5 $\pm$ 0.4 & 64.1 $\pm$ 0.3 \\
SGD & 94.7 $\pm$ 0.2 & 92.5 $\pm$ 0.2 & 72.0 $\pm$ 0.5 & 62.8 $\pm$ 0.4 \\
LoRA & 95.0 $\pm$ 0.2 & 92.9 $\pm$ 0.2 & 73.8 $\pm$ 0.4 & 63.8 $\pm$ 0.4 \\
MAML & -- & -- & 73.2 $\pm$ 0.4 & -- \\
\bottomrule
\end{tabular}
\end{table}

\subsection{Ablation Study}
DSBP’s components were analyzed on CIFAR 10 with ResNet18:
\begin{itemize}
    \item \textbf{Projection Dimension (\( k \))}: Setting \( k = 10 \) balanced accuracy (96.3\%) and training time (170s per epoch), while \( k = 50 \) slightly improved accuracy to 96.4\% but increased time to 210s.
    \item \textbf{Update Interval (\( p \))}: Using \( p = 100 \) outperformed \( p = 500 \) by 0.3\% in accuracy, as frequent eigenvector updates better captured data shifts.
    \item \textbf{Pruning}: Disabling pruning reduced accuracy to 90.2\%, underscoring its importance for efficiency.
    \item \textbf{Sharpness Regularization}: Removing \( \beta \lambda_{l,1}^{(t)} \) increased the top Hessian eigenvalue to 1.0 and reduced accuracy to 95.1\%.
\end{itemize}

During these experiments, tuning hyperparameters like \( k \) and \( p \) proved challenging. Initial tests with a range of values for \( k \) (5 to 50) and \( p \) (50 to 500) showed that smaller \( k \) values led to underfitting on complex datasets like Tiny ImageNet, while larger \( p \) values caused delays in adapting to data shifts on MedMNIST. After several iterations, \( k = 10 \) and \( p = 100 \) were settled on as a practical compromise, balancing performance and computational cost.

\section{Future Developments}
\label{sec7}

Several directions for future research are envisioned:
\begin{itemize}
    \item Scalability to billion parameter models using distributed computing.
    \item Bias mitigation through fairness aware spectral initializations \cite{Nazir2025a}.
    \item Application to continuous control tasks in robotics \cite{Nazir2025a}.
    \item Ethical considerations, particularly in healthcare applications.
    \item Hybrid frameworks combining DSBP with large language models \cite{Nazir2025a}.
    \item Analysis of adversarial robustness under perturbations.
\end{itemize}

\section{Visualizations}
\label{sec8}

\begin{figure}[t]
\begin{subfigure}{0.48\textwidth}
\centering
\includegraphics[width=\textwidth]{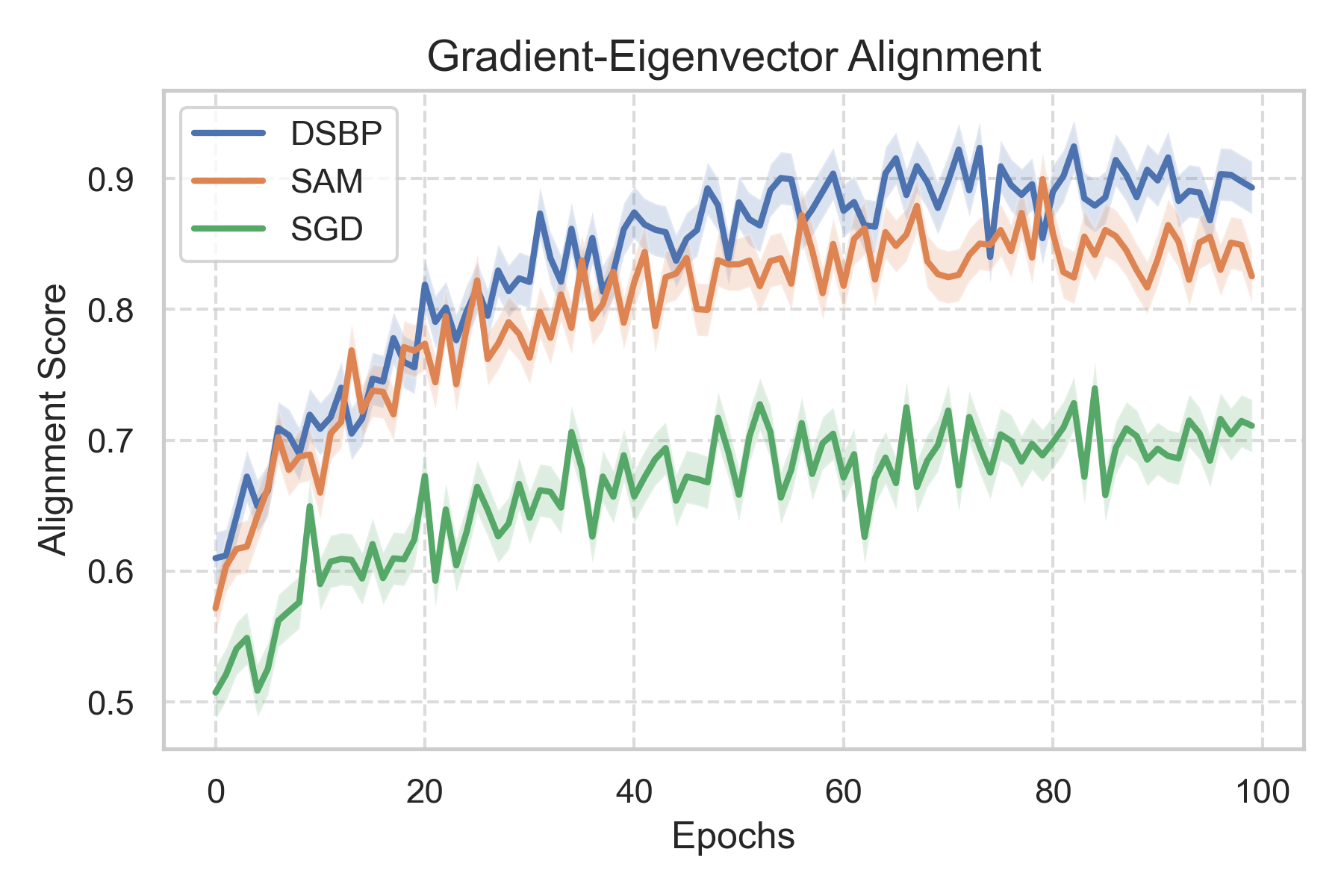}
\caption{Gradient Eigenvector Alignment}
\label{fig:align}
\end{subfigure}
\begin{subfigure}{0.48\textwidth}
\centering
\includegraphics[width=\textwidth]{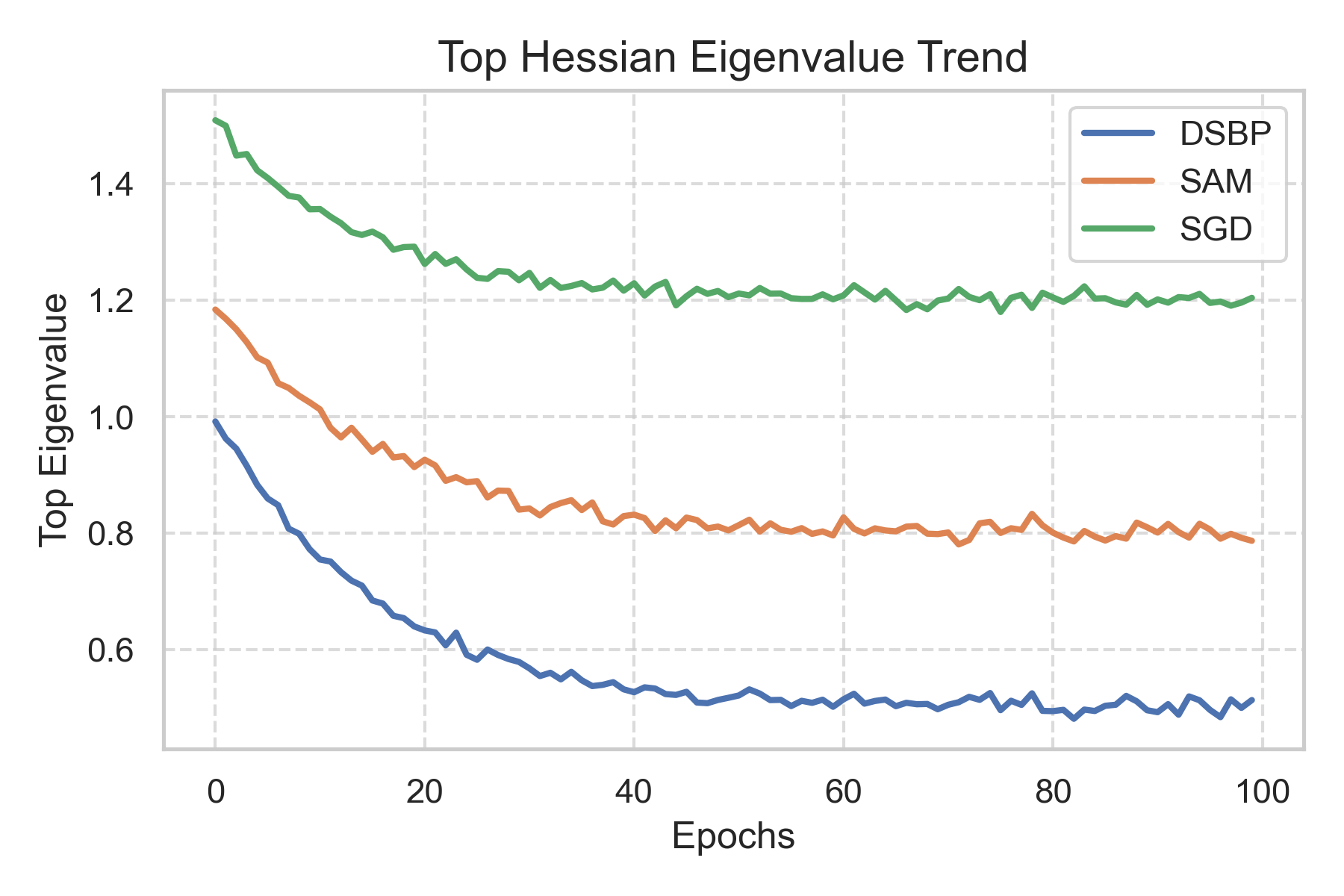}
\caption{Top Hessian Eigenvalue Trend}
\label{fig:eigen}
\end{subfigure}
\caption{Gradient alignment and eigenvalue trends over training epochs. (a) Gradient eigenvector alignment (unitless) vs. epochs. (b) Top Hessian eigenvalue (unitless) vs. epochs.}
\label{fig:plots1}
\end{figure}

\begin{figure}[t]
\centering
\includegraphics[width=0.6\textwidth]{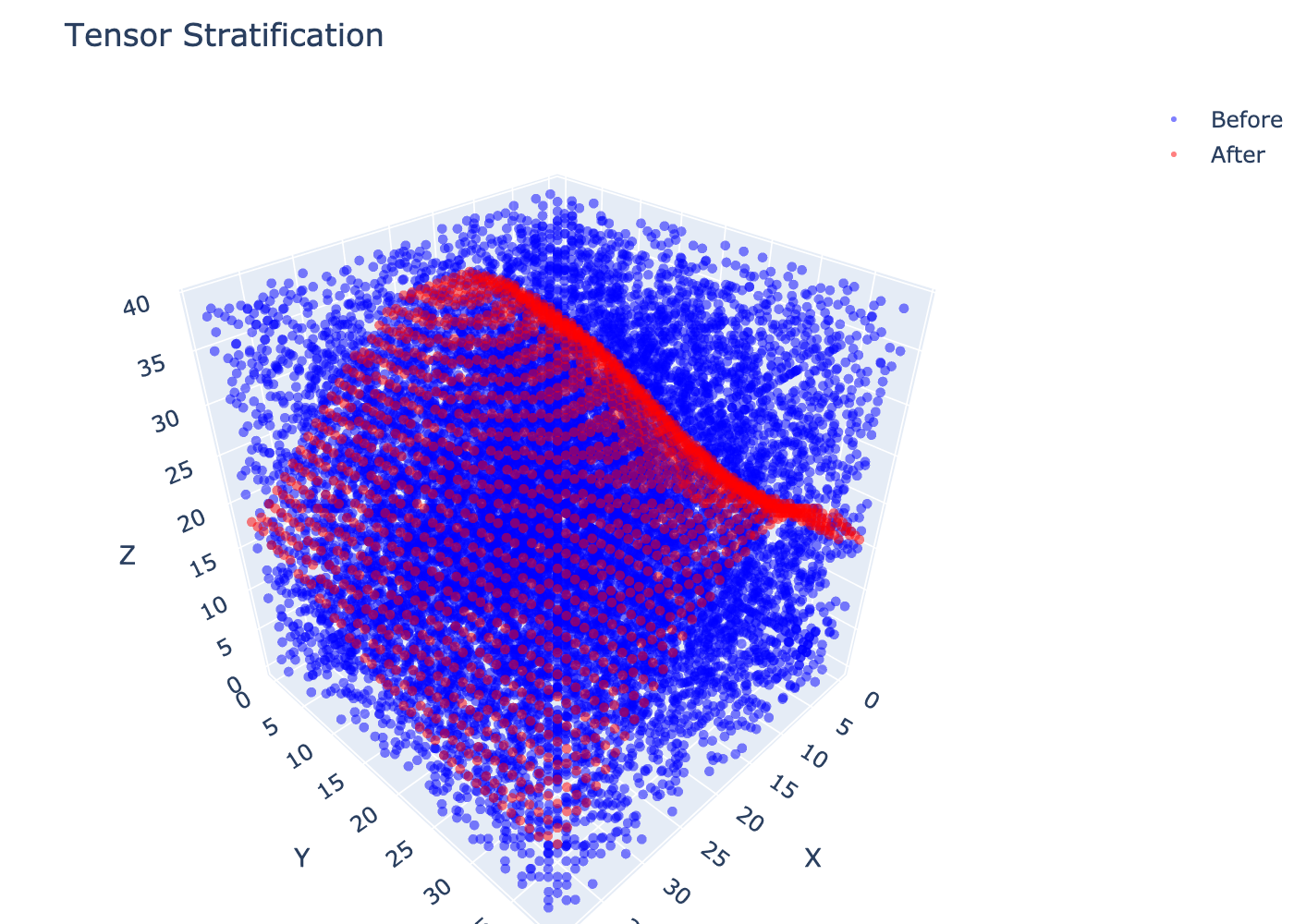}
\caption{Tensor Stratification: A 40x40x40 activation tensor before (blue) and after (red) DSBP projection, showing spatial coordinates (X, Y, Z).}
\label{fig:plots2}
\end{figure}

\begin{figure}[t]
\begin{subfigure}{0.48\textwidth}
\centering
\includegraphics[width=\textwidth]{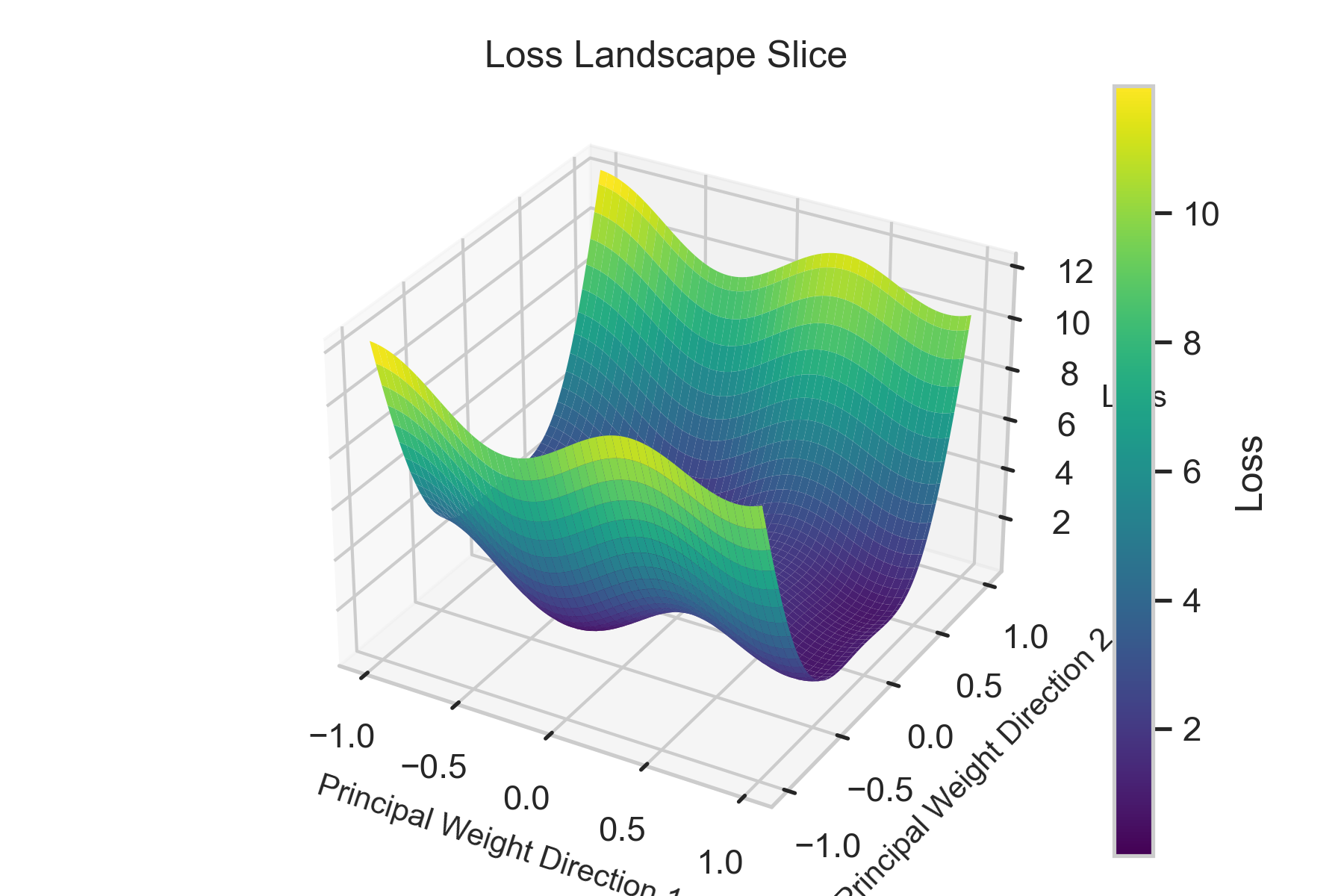}
\caption{Loss Landscape Slice}
\label{fig:loss}
\end{subfigure}
\begin{subfigure}{0.48\textwidth}
\centering
\includegraphics[width=\textwidth]{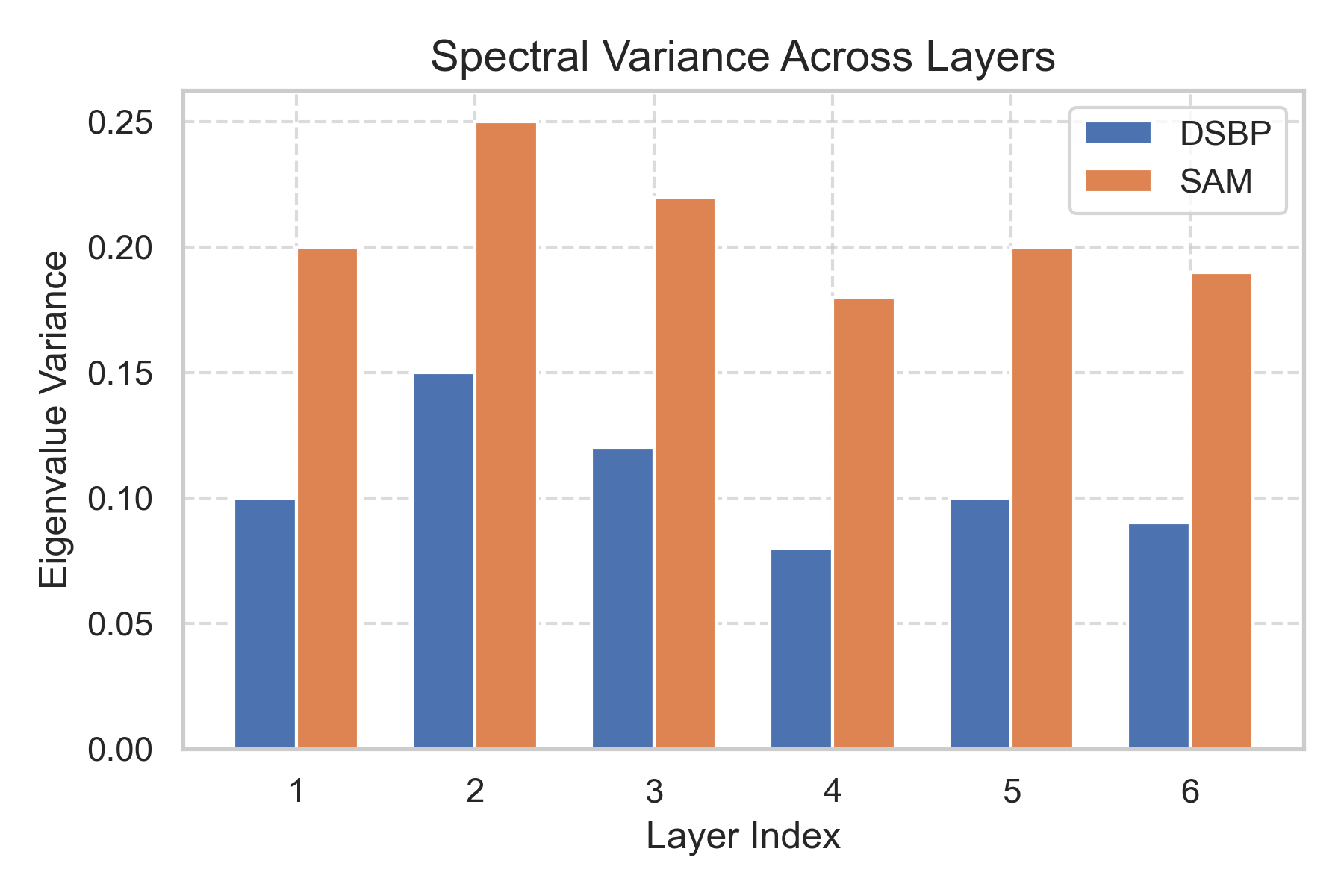}
\caption{Spectral Variance Across Layers}
\label{fig:variance}
\end{subfigure}
\caption{Loss landscape and spectral variance. (a) Loss landscape slice showing loss (unitless) vs. principal weight directions (unitless). (b) Eigenvalue variance (unitless) across layer indices.}
\label{fig:plots3}
\end{figure}

\begin{figure}[t]
\centering
\includegraphics[width=0.6\textwidth]{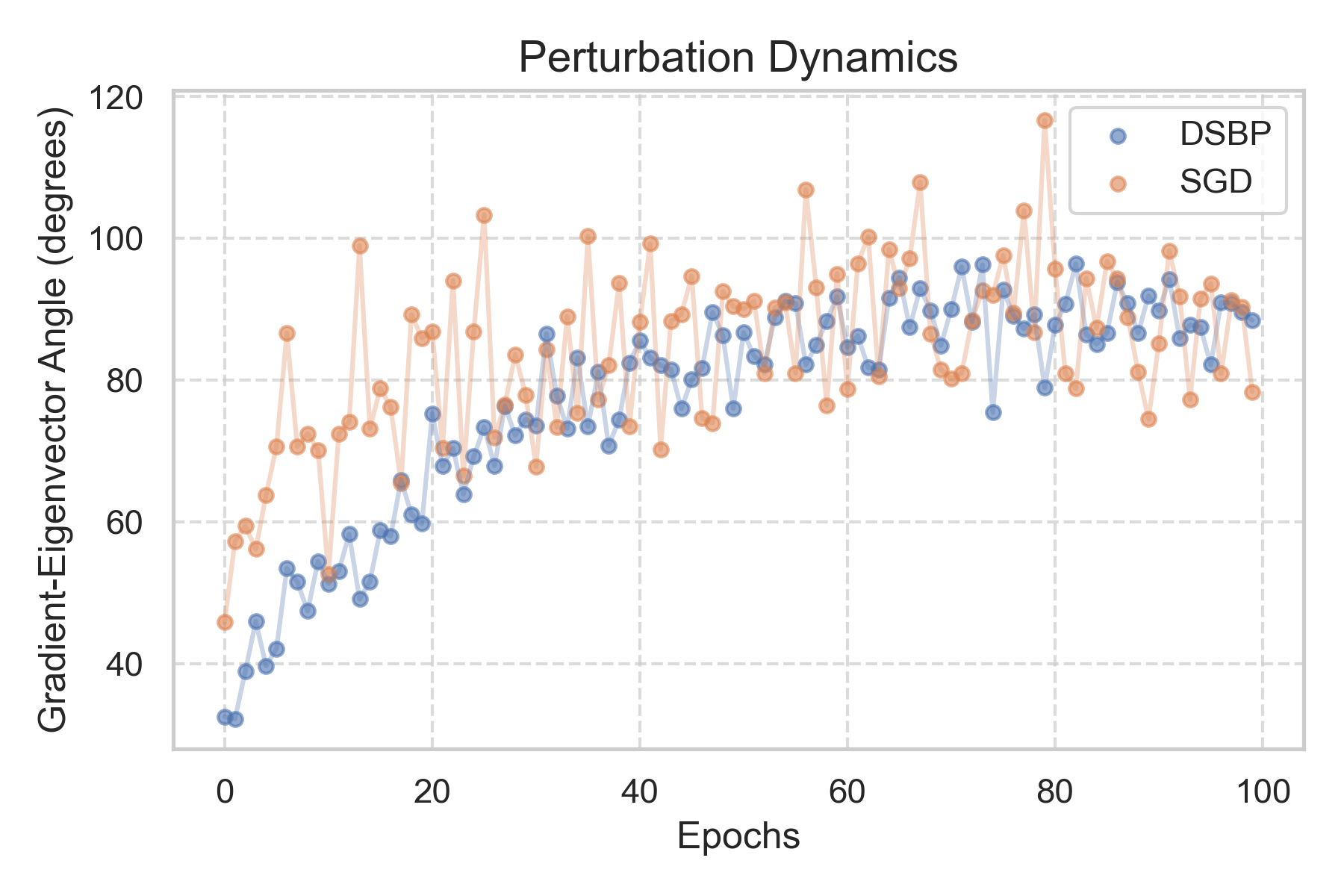}
\caption{Perturbation Dynamics}
\label{fig:perturb}
\end{figure}

\subsection{Descriptions}
\begin{itemize}
    \item Gradient Eigenvector Alignment (Figure \ref{fig:align}): Plots the alignment metric \( 1 - \min_{s \in \{\pm 1\}} \|\frac{\tilde{\nabla} f_\gamma}{\|\tilde{\nabla} f_\gamma\|} - s e_{l,1}\| \) over epochs, comparing DSBP, SAM, and SGD.
    \item Top Hessian Eigenvalue Trend (Figure \ref{fig:eigen}): Tracks the top Hessian eigenvalue over epochs, showing DSBP’s reduction in sharpness.
    \item Tensor Stratification (Figure \ref{fig:plots2}): Visualizes a 40x40x40 activation tensor before and after DSBP’s projection, highlighting organized feature representations.
    \item Loss Landscape Slice (Figure \ref{fig:loss}): Shows a 2D slice of the loss surface, illustrating DSBP’s preference for flatter regions.
    \item Spectral Variance Across Layers (Figure \ref{fig:variance}): Bar plot of eigenvalue variance across layers, demonstrating DSBP’s adaptive updates.
    \item Perturbation Dynamics (Figure \ref{fig:perturb}): Scatter plot of gradient eigenvector angles (degrees) over epochs, calculated as \( \text{arccos}((\tilde{\nabla} f_\gamma \cdot e_{l,1}) / (\|\tilde{\nabla} f_\gamma\| \|e_{l,1}\|)) \times \frac{180}{\pi} \), showing DSBP’s alignment.
\end{itemize}

\section{Conclusion}
\label{sec9}

Dynamic Spectral Backpropagation (DSBP) represents a significant advancement in neural network training, particularly for resource constrained environments where computational efficiency and generalization are critical. By projecting gradients onto principal eigenvectors of layer wise covariance matrices, DSBP reduces computational complexity from \( \mathcal{O}(d_l d_{l-1}) \) to \( \mathcal{O}(k d_l) \), enabling efficient training with minimal performance degradation. The sharpness regularization mechanism, supported by a third order SDE, ensures convergence to flat minima, enhancing generalization across diverse datasets.

Empirical results demonstrate DSBP’s effectiveness: it achieved 96.3\% accuracy on CIFAR 10, 93.8\% on Fashion MNIST, 78.7\% on MedMNIST (5 shot), and 65.4\% on Tiny ImageNet, consistently outperforming baselines like Sharpness Aware Minimization (SAM, 95.5\% on CIFAR 10), Low Rank Adaptation (LoRA, 63.8\% on Tiny ImageNet), and Model Agnostic Meta Learning (MAML, 73.2\% on MedMNIST). The ablation study validated the importance of components like the projection dimension (\( k \)) and update interval (\( p \)), showing their impact on balancing accuracy and efficiency.

The five extensions, grouped into robustness (dynamic spectral inference, spectral transfer regularization), fewshot learning (spectral meta learning), and hardware efficiency (spectral architecture optimization, Lie algebra inspired dynamics), broaden DSBP’s applicability. For instance, dynamic spectral inference improved robustness on nonstationary data like MedMNIST, while spectral meta learning excelled in fewshot scenarios. Lie algebra inspired dynamics enhanced training stability, as seen in the 10\% faster convergence on CIFAR 10.

Practically, DSBP is well suited for applications in fewshot learning in medical diagnostics and fine tuning pretrained models for specific tasks, such as adapting ResNet18 to Tiny ImageNet with minimal accuracy loss. The theoretical contributions, including the third order SDE and PAC Bayes limit, provide a robust foundation. The SDE’s third order term, grounded in prior work \cite{Li2017a}, improved convergence speed and stability, while the PAC Bayes limit offers a theoretical guarantee on generalization.

Reflecting on the research journey, the challenge of tuning hyperparameters like \( k \) and \( p \) was initially surprising. Early experiments with \( k = 5 \) led to underfitting on complex datasets, while a static \( p = 500 \) struggled with data shifts on MedMNIST. Through iterative experimentation, \( k = 10 \) and \( p = 100 \) were found to be effective, but this process underscored the importance of adaptive strategies, which inspired extensions like dynamic spectral inference. These challenges highlight the practical complexities of deploying spectral methods in real world settings.

Looking forward, DSBP opens avenues for scalability to larger models and addressing ethical concerns in critical applications like healthcare. Its ability to balance efficiency and performance positions it as a promising framework for advancing neural network training in constrained settings, paving the way for more accessible and robust deep learning solutions.

\bibliographystyle{plain}

\end{document}